%% file: main.tex
\documentclass[10pt,journal,compsoc]{IEEEtran}

\usepackage[noadjust]{cite}
\usepackage{filecontents}

\ifCLASSINFOpdf
\else
\fi

\usepackage{graphics}
\usepackage{lipsum}
\usepackage{graphicx,amsmath,alltt}
\usepackage[figure,linesnumbered]{algorithm2e}
\usepackage{textcomp}
\usepackage{booktabs}
\usepackage{multirow}
\usepackage{colortbl}
\usepackage{ragged2e}
\usepackage{tabularx,booktabs}
\usepackage[dvipsnames,table,xcdraw]{xcolor}
\renewcommand{\citedash}{--} 

\begin{document}


\title{Handcrafted Feature Selection Techniques for Pattern Recognition: A Survey}

\author{
    \IEEEauthorblockN {
       Alysson~Ribeiro~da~Silva$^{*}$,
       and Camila~Guedes~Silveira$^{*}$
    }
   
    \IEEEauthorblockA {
        $^{*}$Computer Science Graduate Program \\ 
        Federal University of Minas Gerais \\
        Belo Horizonte, Brazil
    }
}

\IEEEtitleabstractindextext{%


\justify
\begin{abstract}
The accuracy of a classifier, when performing Pattern recognition, is mostly tied to the quality and representativeness of the input feature vector. Feature Selection is a process that allows for representing information properly and may increase the accuracy of a classifier. This process is responsible for finding the best possible features, thus allowing us to identify to which class a pattern belongs. Feature selection methods can be categorized as Filters, Wrappers, and Embed. This paper presents a survey on some Filters and Wrapper methods for handcrafted feature selection. Some discussions, with regard to the data structure, processing time, and ability to well represent a feature vector, are also provided in order to explicitly show how appropriate some methods are in order to perform feature selection. Therefore, the presented feature selection methods can be accurate and efficient if applied considering their positives and negatives, finding which one fits best the problem\textquotesingle s domain may be the hardest task.
\end{abstract}


\begin{IEEEkeywords}
Feature Selection; Pattern Recognition; Filter; Wrapper; Computer Vision; Machine Learning
\end{IEEEkeywords}}

\maketitle
\IEEEdisplaynontitleabstractindextext
\IEEEpeerreviewmaketitle
\ifCLASSOPTIONcompsoc


\input{tex/1_introduction}
\input{tex/2_techniques}
\input{tex/3_filter}
\input{tex/4_wrapper}
\input{tex/5_conclusion}


\bibliographystyle{IEEEtran}
\bibliography{main}

\end{document}

%% file: tex/1_introduction.tex
\section{Introduction}

In machine learning, representing and organizing data into meaningful information is a fundamental key that allows a classifier, such as neural networks, to determine the belongingness of observations into classes \cite{2014_Feature_extraction_electroencephalogram_signal_review}. Observed data can be obtained from sources, in different domains, such as robot sensors, digital cameras, medical instruments, and digital games agents, where the main objective is to allow a computer program to perform decision-making \cite{2017_DeepLearning_generalAI}. Organize and representing data is not a trivial task, where the main problem is the lack of a standard and fast way to evaluate and select a minimal set of information to represent an observed object that will maximize a classifier\textquotesingle s accuracy \cite{2011_Feature_selection_methods_algorithms}. In order to tackle that problem, feature selection techniques are used, where meaningful information, obtained from observed objects on the task\textquotesingle s domain is selected and stored inside feature vectors used as a classifier\textquotesingle s input.

Feature selection methods use search algorithms or heuristics that seek for a set of features, from a feature vector, that when evaluated, all together, through an evaluation function $f$ will maximize its value by reaching a local or global optimum. The search space, where feature selection occurs, is composed of all possible combinations of features that compose a feature vector, thus the selection process may not be able to maximize $f$ properly. In order to handle a search space, trying to maximize $f$, a feature selection technique can be built upon three categories \cite{2011_Feature_selection_methods_algorithms} described as follows:

\begin{enumerate}
	\item The first category is called \textit{Filter}, where it is used to describe feature selection methods not bound to a classifier.
	\item A second category is called \textit{Wrapper} and it is responsible to describe methods that rely, for the most part, on a classifier.
	\item The third category, called \textit{Embed}, describes hybrid approaches (combination of \textit{Filters} and \textit{Wrappers}).
\end{enumerate}

This paper presents a survey on some \textit{Filters} and \textit{Wrappers} methods for handcrafted feature selection since \textit{Embed} ones are hybrids. Some discussions, with regard to the data structure, processing time, and ability to well represent a feature vector, are also provided in order to explicitly show how appropriate some methods are in order to perform feature selection.

The rest of this document is organized as Follows. In Section 2, an overview of pattern recognition is given addressing the importance of feature selection during the process. Furthermore, Section 3 presents some relevant feature selection techniques related to \textit{Filter} methods. In addition, Section 4 addresses \textit{Wrappers}, where most evolutionary algorithms are presented as feature selection tools. Next, in Section 5 a discussion on the advantages and disadvantages of each addressed method is presented. Finally, in Section 6, conclusions of the performed research are presented.

%% file: tex/2_techniques.tex
\section{Pattern Recognition Process}

The pattern recognition process allows for the recognition of classes of objects from an observed sample. The classifier's role is to decide in which class a received external feature vector corresponds. By performing this task, it is possible to identify objects in reality if they are represented through feature vectors. The main problem of this process is that raw data obtained from reality can\textquotesingle t be used directly by a classifier since it has noise and doesn\textquotesingle t contain much information that can be used to compare among different classes. In order to tackle that problem, the classification process often routines such as:

\begin{itemize}
	\item \textbf{Data description}: used to describe and organize raw data obtained from reality into meaningful information, through various descriptors, that can be used properly by a classifier.
	\item \textbf{Information assembler}: allows to assembler of all the used descriptors in order to form feature vectors that a classifier can handle.
	\item \textbf{Feature selection}: The final stage of the process, in which only meaningful information, that maximizes the classifier\textquotesingle s accuracy is selected to compose its input vectors.
\end{itemize}

\begin{figure*}[th]
	\begin{center} 
		\includegraphics[width=1.0\textwidth,height=!]{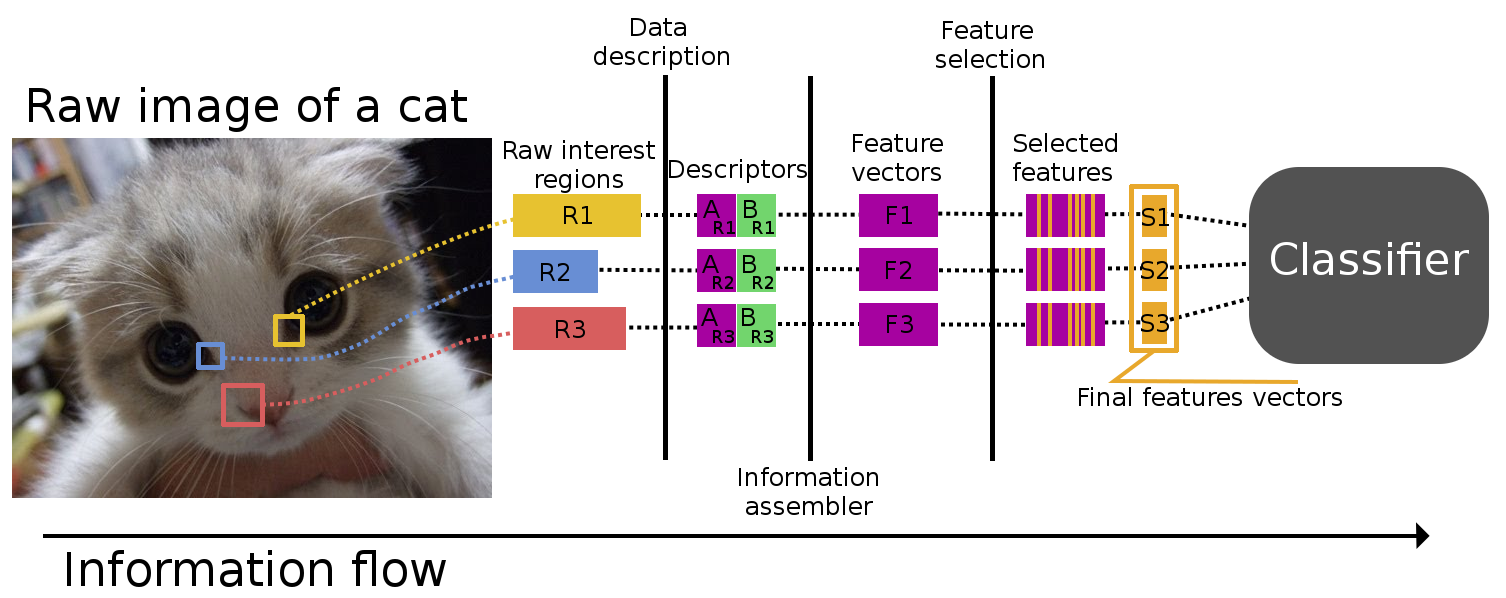}
	\end{center} 
	\vspace{-10px}
	\caption{Image Pattern recognition process example, from interest region detection to feature selection. The raw image of a cat, on the left side of the figure, is an example image given for a classifier in order to be classified. However, its raw pixels can not be introduced directly inside of a classifier. To tackle that, the \textit{Data description} step operates under the presented interest region data stored inside $R1$, $R2$ and $R3$, where through the descriptors $A$ and $B$, generates the tuples $(A_{r1},A_{r1})$, $(A_{r2},A_{r2})$ and $(A_{r3},A_{r3})$ for classification. On the other hand, the \textit{Information assembler} step, generates the feature vectors $F1$, $F2$, and $F3$ by fusing the descriptors $A$ and $B$ obtained information. At the final step of the process, the meaningful features, from $F1$, $F2$, and $F3$, are selected in order to compose $S1$, $S2$, and $S3$ that can be used by a classifier.}

	\label{fig:featureSel}
	\line(1,0){250}
	\vspace{-10px}
\end{figure*}

A classifier can use as many descriptors, to describe data, and features as it wants, thus allowing it to decide what will benefit itself. As shown by the example in Fig. \ref{fig:featureSel}, of pattern recognition in image processing, the classification of a received image occurs after detecting interest regions, transforming their data into meaningful information, through the descriptors $A$ and $B$, and finally selecting appropriate features that will lead to the classifier\textquotesingle s input vectors $S1$, $S2$ and $S3$.

In the subsequent sections of this paper some feature selection techniques, that help in maximizing a classifier\textquotesingle s accuracy, are presented. In order to understand some of the proposals, a feature vector mathematical description is given as follows. A feature vector $F$ is equal to $\{f_1,...,f_n\}$, where each $f_i \in F$ is a feature that represents information obtained from a descriptor and $n$ is the total amount of features that compose the feature vector $F$. It is important to note that $F$ is composed of one or many descriptor vectors, as shown by the example in Fig. \ref{fig:featureSel}, thus it can be feasible in applying some feature selection techniques to descriptor vectors separately.

%% file: tex/3_filter.tex
\section{Filter approaches}

This Section presents the feature extraction methods based on filter approaches, were given a feature vector $E$, the Filter approach will evaluate $E$ through statistical measurements not bounded to a classifier, such as the correlation between features, and then generates a subset $S$ composed of the most correlated features from $E$.

\subsection{Feature threshold technique}
\label{sec:thresh}

The feature threshold is the simplest filter technique that allows selecting a subset of $F$. It works by calculating the variance of each feature inside an observed population of feature vectors. Each variance is stored inside a feature vector $\bar{F} = \{\bar{f_1},...,\bar{f_n}\}$, where each $\bar{f_1} \in \bar{F}$ is the variance of the feature $f_1$ observed by sampling the feature vector $F$. In order to select features, a threshold value $t$ is defined. The selected feature set $S$ is composed of all features, from $F$, that are above $t$. The main idea of the threshold is indicating how that feature value is variating through all samples, thus allowing us to perform an inference that will lead to concluding about its importance on the final feature set $S$. This method is fast to be performed on a high dimensional feature vector $F$. However, the features that it selects are not correlated, because the variance of each one is checked separately. In conclusion, this method can throw away features that may have some relevance to the classification process.

\subsection{Feature selection through Euclidean Distance}
\label{sec:euclid}

In order to allow verifying the correlation between features, Euclidean distance can be used instead \cite{khalid2014survey}. It is a common way in measuring how far are two observed values. When using this metric for selecting features, firstly a sample vector $S$ need to be computed, where $S = \{s_1, ..., s_n\}$, being each $s_i \in S$ the mean, or expected value, from observations of the feature $f_i$ obtained from several samples. With the newly computed feature vector $S$, the euclidean factor is calculated through Equation \ref{eq:dist},

\begin{equation}
\label{eq:dist}
\texttt{$d_k = \sum_{i}^n (s_k - s_i)^2$}
\end{equation}

\noindent
where $s_k$ is the feature that will be tested and $s_i$ the desired feature $i$ that will be compared with the feature $k$. The features with the highest calculated $d_k$ will be selected based on a threshold value $t$, where if $d_k > t$, then the feature $f_k \in F$ will be selected. The Euclidean distance method solves the lack of relationship between features, as argued in Section \ref{sec:thresh}. Furthermore, it adds extra complexity when computing the feature relevance through $d_k$. However, it still uses a threshold value that can throw away relevant features according to the mean feature vector $S$. Also, it does not consider the class relevance during the classification process when using the computed feature vector $S$, thus the selected features may not represent well all the classes that are handled by a classifier.

\subsection{Selecting features through a $\chi^2$ hypothesis test}

As argued in Section \ref{sec:euclid}, the distance can not be used alone since it does not consider the class/feature relationship. Furthermore, it does not seems feasible to determine features by a threshold value, since it can throw away relevant features. In order to tackle that, the $\chi^2$ test can be used instead, where it is a statistical measurement, used by a hypothesis test, that allows calculating correlation between observed and expected values without a threshold $t$ \cite{vafaie1994feature}\cite{khalid2014survey}\cite{jin2006machine}. In feature selection, the $\chi^2$ test is performed to select a feature from $F$ through Equation \ref{eq:xi},

\begin{equation}
\label{eq:xi}
\texttt{$\chi^2_{f_k \in F} = \sum_{i}^r \sum_{j}^c {(O_{ij} - E_{ij})^2 \over E_{ij}}$}
\end{equation}

\noindent
where $r$ is all the observed values for a feature $f_k \in F$, $c$ all possible classes that the classifier was trained to classify, $O_{ij}$ is the total amount of occurrences that the feature $f_k \in F$ received the value $i$ and a classifier identified the full feature vector $F$ as belonging to a class $j$, $E_{ij}$ is the expected value computed as $(p \times q) \over m$, where $p$ is the total amount of observations that received the value represented by $i$, $q$ the total amount of times that the value represented by $i$ was identified as belonging to the class $j$ and $m$ the total amount of observations. After calculating the $\chi^2$ value of a feature $f_k$, then a hypothesis test, based on a $\chi^2$ distribution, is performed with $\chi^2_{f_k}$, considering a significance level $\alpha$, in order to decide if the feature $f_k$ will be discarded or not. The  $\chi^2$ test considers the influence of the class on the expected value of a feature across multiple samples. However, it is bounded to the convergence of features in reality, where can not be guaranteed that the feature will behave as expected, thus it does not deal well with anomalies.

\subsection{Correlation feature selection}

The correlation feature selection method deals with the feature-to-feature and feature-to-class correlation without performing a hypothesis test, thus it does not rely on expected values. It was proposed by \cite{1999_Correlation_based_feature_selection_machine_learning} and it calculates the correlation level, $M_s$, of a subset $S \subset F$ to a class $C$. It considers also the correlation between each element of $S$, where the subset with the max correlation with $C$ and the minimal correlation between itself is the selected feature set. This method is deployed with an algorithm that handles Equation \ref{eq:cbf},

\begin{equation}
\label{eq:cbf}
\texttt{$M_s = {k \bar{r_{cf}} \over \sqrt{k + k(k-1)\bar{r_{ff}}}}$}
\end{equation}

\noindent
where $M_s$ is the evaluated feature subset, $k$ is the total amount of features, $\bar{r_{cf}}$ is the average correlation between $M_s$ and the class $C$, calculated for each feature, and $\bar{r_{ff}}$ is an average correlation between each $f_i \in M_s$. According to \cite{1999_Correlation_based_feature_selection_machine_learning}, the numerator of Equation \ref{eq:cbf} represents how good is the subset $S$ to identify the class $C$ and the denominator shows a total amount of redundancy between al features inside $S$. This approach can deal well by correlating features with class and features with features. However, it still does not considers how the classifier behaves, and its accuracy when classifying, thus it may select features that do not improve, in a substantial way, the quality of the classification process. In order to tackle that problem, \textit{Wrapper} approaches can be used instead.

%% file: tex/4_wrapper.tex
\section{Wrapper approaches}

In this Section, we present wrapper approaches. For instance, a wrapper behavior will evaluate a subset $S$ of a feature vector $E$, where the value of $f$ is the output of a classifier. If the subset $S$ satisfies an evaluation criterion, such as a threshold value $t$, then the searching process will stop and $S$ will be the selected feature vector.

\subsection{Sequential Forward Selection}

In contrast to \textit{Filters}, \textit{Wrapper} methods try to achieve maximum representativeness by interacting directly with the classifier. A simple way in achieving that interaction is by using the Sequential Forward Selection (SFS) method \cite{mao2004orthogonal}\cite{2011Featureselectionmethodsalgorithms}\cite{ khalid2014survey}. It is an algorithm bottom-up, from features to classifier, where a set $S$, made by features, is built according to improvements made on the classification process \cite{mao2004orthogonal,2011Featureselectionmethodsalgorithms}. This method can be assembled in various ways, and in this paper, only a greedy version of it is provided. As shown by Fig. \ref{alg:greedyforward}, the Sequential Forward Selection receives two parameters, where $g$ is the evaluation function or a classifier\textquotesingle s output, and a universe feature set $U$, that contains all features that will be tested and an input set $\Omega$ used to start the searching process.

\begin{algorithm}[t]
	\SetKwInOut{Input}{input}
	\SetKwInOut{Output}{output}
	\Input{Evaluation function $g$, feature set $U$, input set $\Omega$}
	\Output{Selected features set $S$}
	\BlankLine
	\DontPrintSemicolon
	\Begin{
		$S \gets \Omega$\;
		$g_{old} = g(S)$\;
		\For{\textbf{each} $f_i \in U$}
		{
			$Temp \gets S + \{f_i\}$\;
			$g_{new} = g(Temp)$\;
			\If{$g_{new} > g_{old}$}
			{
				$S \gets S + \{f_i\}$\;	
				$g_{old} = g_{new}$\;
			}
		}

		Return $S$\;
	}
	
	\vspace{-10px}
	\caption{Greedy Sequential Forward Selection (SFS) pseudo-code.}
	\label{alg:greedyforward}
	\vspace{-10px}
\end{algorithm} 

The algorithm, in Fig. \ref{alg:greedyforward}, firstly initialize the selected feature $S$ with an empty set $\emptyset$ and calculates, through $g$, the value $f_{old}$ that the classifier gives to it. Next, for each feature $f_i \in U$ it creates a construction called $Temp$, which is used to calculate the influence of inserting the feature $f_i$ into $S$. The newly formed construction $Temp$ represents a variation of $S$ and it is evaluated through $g$, where the evaluation results are stored inside $g_{new}$. If the feature $f_i$ shows a positive change obeying the inequality $g_{new} > g_{old}$, then the feature $f_i$ is persisted into $S$. The variables $g_{old}$ and $g_{new}$ are used to verify the $\delta g$ change in the quality of the generated solution inside $S$ so far. The algorithm ends by returning the vector $S$ that contains the selected features. This algorithm has two main flaws; firstly, it does not consider the correlation between all possible combinations of features inside $F$; secondly, it must search each feature in sequential order, thus the execution time of the algorithm rely on the number of features inside the $F$.

\subsection{Sequential Backward Elimination}

Another method based on successive tests inside a feature vector is the Sequential Backward Elimination (SBE). Differently from the SFS, the SBE searches for all possible combinations inside $F$. As shown by Fig. \ref{alg:greedybackward}, this method starts with a full feature vector $U$ and for each $f_i \in U$ it selects, by a brute force algorithm, the one in which the removal caused a minor change on an observed function $g$ \cite{mao2004orthogonal}\cite{2011Featureselectionmethodsalgorithms}\cite{ khalid2014survey}.

\begin{algorithm}[t]
	\SetKwInOut{Input}{input}
	\SetKwInOut{Output}{output}
	\Input{Evaluation function $g$, feature set $U$, iteration limit $t$}
	\Output{Selected features set $S$}
	\BlankLine
	\DontPrintSemicolon
	\Begin{
		$S \gets U$\;
		$f_{remove} = null$\;
		\While{$t \ge 0$}
		{
			$g_{max} = 0$\;
			\For{\textbf{each} $f_i \in U$}{
				$Temp \gets S$\;
				$Temp = Temp - \{f_i\}$\;
				$g_{new} = g(Temp)$\;
				\If{$g_{new} > g_{max}$}{
					$f_{remove} = f_i$\;
					$g_{max} = g_{new}$\;	
				}
			}
			$S = S - \{f_{remove}\}$\;
			$t = t - 1$\;
		}
		
		Return $S$\;
	}
	
	\vspace{-10px}
	\caption{Greedy Sequential Forward Elimination (SBE) pseudo-code.}
	\label{alg:greedybackward}
	\vspace{-10px}
\end{algorithm} 

The algorithm, depicted in Fig. \ref{alg:greedybackward}, receives an extra parameter, in comparison to the one depicted in Fig. \ref{alg:greedyforward}, called $t$. This parameter is responsible to tell the algorithm when to stop selecting features to be removed from $U$. The algorithm starts by configuring $S$ as the feature universe set $U$. For each iteration $t$, it will find the feature $f_i \in U$ to be removed. The $Temp$ set is used to perform operations guaranteeing the integrity of $S$. In order to find which $f_i$ to remove, for each $f_i$ it will remove it from $Temp$ and compute $g_{new} = g(Temp)$. If the value stored inside $g_{new}$ is the greatest among all other global maximum stored inside $g_{max}$, then it is called $g_{max}$ and the feature $f_i$ is stored inside $f_{remove}$ for removal. It is important to note that the structure of $Temp$ will always equal $S$, even after removing a feature from $Temp$, until $S$ gets updated at the end of one iteration. Next, the algorithm updates the set $S$ by doing $S = S - \{f_{remove}\}$, thus the remaining features inside $S$ are the ones selected by the algorithm. Finally, it returns $S$ in order to be used by a classifier. This method seems more feasible to select features since it checks all possible combinations inside $F$. However, it has a computational time equal to $n^2$, where $n$ is the total amount of features inside $F$, thus the method seems unfeasible if $F$ posses an arbitrarily high amount of features. 

\subsection{Plus-L Minus-R Selection}

In order to reduce the computational time of the SFS, the Plus-L Minus-R Selection (LRS) can be used instead. The LRS is a heuristic that works on a visibility window and it is based on the (SFS) and (SFE), where features are added and removed according to a search criteria. The SLR algorithm, depicted in Fig. \ref{alg:lrs}, builds a feature set $S$ from a feature universe $U$ by adding $l$ features and removing $r$ features until reaching a stopping criteria \cite{khalid2014survey}\cite{ mao2004orthogonal}.

\begin{algorithm}[t]
	\SetKwInOut{Input}{input}
	\SetKwInOut{Output}{output}
	\Input{Evaluation function $g$, feature set $U$, iteration limit $t$}
	\Output{Selected features set $S$}
	\BlankLine
	\DontPrintSemicolon
	\Begin{
		$l = rand()$ // random number between 0 and feature set size \;
		$r = rand()$ // random number between 1 and feature set size \;
		$S \gets U$\;
			
		\If{$l > r$}{
			$S \gets \emptyset$\;
		}
	
		\While{$t > 0$}{
			\For{$i=0;i<l;i=i+1$}{
				$S = S + \{max(g,U)\}$\;
			}
			\For{$i=0;i<r;i=i+1$}{
				$S = S - \{min(g,U)\}$\;
			}
			$U = U - S$\;
			$t = t - 1$\;		
		}
	
		return $S$\;
	}
	
	\vspace{-10px}
	\caption{Plus-L Minus-R Selection pseudo-code.}
	\label{alg:lrs}
	\vspace{-10px}
\end{algorithm} 

The heuristic in Fig. \ref{alg:lrs}, receives an evaluation function $g$, used to evaluate the partially generated solution inside $S$, a feature universe $U$, containing all features that will be used in the search process, and an iteration limit $t$. It first initializes the values of $l$ and $r$ with random variables that will range from $0$ to the total amount of features inside $U$. Then, if $l > r$, the working set will be initialized as $S \gets \emptyset$. The main loop of the algorithm will add $l$ features, selected through the $max$ function, and remove $r$ features, selected through the $min$ function, inside $S$. The $max$ function selects the feature that when added to $S$ caused the maximum amount of change when computing $g$, by the other hand, the $min$ function will select the feature that when removed from $S$ caused a minimal amount of change when computing $g$. At the end of the process the possible features to select, inside $U$, are updated by $U = U - S$. At the end of the algorithm, the selected features are returned inside the feature vector $S$. 

When using the LRS, features are selected through window sampling, thus allowing to avoid a high computational time. The main flaw of the algorithm is related to the fact that the search window starts always at the beginning of $F$, thus for every iteration, it will start a new search procedure from scratch. Furthermore, it can not be guaranteed that a good solution will be reached since the heuristic relies on a stopping criterion $t$.

\subsection{Hill Climbing}

In order to avoid searching inside a window defined by random variables, as accomplished by the LRS, the Hill Climbing algorithm is used to find a local optimum from a given function as described in Equation \ref{eq:hill}, 

\begin{equation}
\label{eq:hill}
\texttt{$f = x_1a_1 + x_2a_2 + ... + x_na_n$}
\end{equation}

\noindent
where its main objective is to adjust each coefficient, $x_i$ from $f$, in order to change its value towards local optima. The Hill Climbing algorithm, depicted in Fig. \ref{alg:hill}, receives a coefficient vector $C$ that contains all the coefficients for an also given function $g$. When selecting features, the coefficient vector $C$ of the Hill Climbing algorithm is composed by $\{c_1,...,c_n\}$, where each $c_i \in C$ is a binary variable used as a coefficient that represents the presence or absence of the feature $i$. \cite{moore1992empirical}

\begin{algorithm}[t]
	\SetKwInOut{Input}{input}
	\SetKwInOut{Output}{output}
	\Input{Evaluation function $g$, function coefficients $C$}
	\Output{Selected features set $S$}
	\BlankLine
	\DontPrintSemicolon
	\Begin{
		$hasChange = true$\;
		\While{$hasChange$}
		{
			$hasChange = false$\;
			$g_{old} = g(C)$\;
			\For{\textbf{each} $x_i \in C$}
			{
				$TempLeft \gets C$; $TempLeft[x_i] = 0$\;
				$TempRight \gets C$; $TempRight[x_i] = 1$\;
				$gLeft_{new} = g(TempLeft)$\;
				$gRight_{new} = g(TempRight)$\;				
				
				\If{$gLeft_{new} > gRight_{new}$ \& $gLeft_{new} > g_{old}$}
				{
					$hasChange = true$; $C \gets TempLeft$; break\;
				}\Else{
					\If{$gRight_{new} > g_{old}$}
					{
						$hasChange = true$; $C \gets TempRight$; break\;
					}
				}				
			}
		}
		return $C$\;
	}
	
	\vspace{-10px}
	\caption{Forward Hill Climbing algorithm pseudo-code.}
	\label{alg:hill}
	\vspace{-10px}
\end{algorithm} 

The Hill Climbing algorithm for feature selection starts by initializing a control variable $hasChange$ to true in order to enter its main loop. Next, it calculates the current value of $g$ based on a coefficient vector $C$. In order to decide what feature to change in order to achieve a better coefficient vector, the algorithm will compute a left step and a right step for each feature $x_i \in C$. In order to calculate the left and right steps, a copy of $C$ is created inside $TempLeft$ and $TempRight$, respectively. The function value of $g$ is computed for each generated step and stored inside $gLeft_{new}$ and $gRight_{new}$. The algorithm then selects the highest change between $gLeft_{new}$ and $gRight_{new}$ and compares it with $g_{old}$, if this calculated delta is positive the algorithm identified a change and assigns that change by copying $TempLeft$ or $TempRight$ into $C$ and stops the iteration. This process will repeat until there are no more possible changes inside $C$. The returned $C$ variable will contain a feature vector, composed of binary variables representing the presence or absence of features, that need to be interpreted in order to achieve feature selection. The main problem of this algorithm is that it can not be guaranteed that it will reach a global optimum since it can not see the entire search space. 

\subsection{Simulated Annealing}

In order to avoid the local optima problem of the Hill Climbing algorithm, the Simulated Annealing (SA) is an algorithm that uses random leaps instead of a single step \cite{lin2008parameter}. The random leaps are controlled by the principle of using a temperature variable that controls the leap length and also to decide when to stop seeking a solution. It works based on states, where each state represents a configuration of the analyzed object that is associated with a value from a behavior function $g$. The SA, when performing feature selection, is configured with a super set of features $sU$, where each $su_i \in sU$ is a possible configuration of a feature set $U$. For example, if the vector $I = \{0,0,0\}$, being $i_i \in I$ a binary variable that represents the presence or absence of the feature $i$, then a possible configuration of $I$ would be $I = \{1,0,0\}$, where the feature $1$ is selected and all others are discarded.

In order to perform the SA and achieve feature selection \cite{lin2008parameter}, the algorithm in Fig. \ref{alg:sa} receives an evaluation function $g$, an initial temperature $T$, an iteration limit $t$, and a superset of feature $sU$. The algorithm starts by selecting the first state of the observed object, randomly, and storing it inside $currentS$. Next, it iterates until reaching the stopping criteria, where a new state $S$ will be checked as a potential new solution. The algorithm checks the new solution by testing if its $P(energy1, energy2, T)$ is higher than a generated random number $\alpha \in$ [$0$,$1$].

\begin{algorithm}[t]
	\SetKwInOut{Input}{input}
	\SetKwInOut{Output}{output}
	\Input{Evaluation function $g$, initial Temperature $T$, iteration limit $t$, super set of features $sU$}
	\Output{Selected features set $S$}
	\BlankLine
	\DontPrintSemicolon
	\Begin{
		$f = t$\;
		$currentS = randState(sU)$\;
		\While{$f \ge 0$}
		{
			$T = updateTemp(f / t)$\;
			$S = select(sU)$\;
			$energy1 = E(S)$\;
			$energy2 = E(currentS)$\;
			\If{$P(energy1, energy2, T) > rand()$}
			{
				$currentS \gets S$\;
			}
			$f = f - 1$\;		
		}	
	
		return $currentS$\;
	}
	
	\vspace{-10px}
	\caption{Simulated Annealing for feature selection pseudo-code.}
	\label{alg:sa}
	\vspace{-10px}
\end{algorithm} 

Both, the $energy1$ and $energy2$ variables, represent the energy of an observed state or the value of $g$ when observing that state. At the end of the algorithm, the $currentS$ is the state that contains the optimal solution that was found during the process. The SA can avoid getting stuck in local optima, however, it needs to store the energy of each state in order to work properly. Consequently, each possible combination of all components of the feature vector $F$ will be stored, thus being unfeasible to handle a large feature vector.

\subsection{Genetic Algorithm}

In order to tackle being stuck in a local optimum and avoid storing each possible combination of the feature vector $F$, the Genetic Algorithm (GA) can be used instead of the SA \cite{khalid2014survey}\cite{vafaie1994feature}\cite{jain1997feature}. It is based on the theory of evolution, where a population of individuals, coded as feature vectors, passes through a process of breeding, birth, and even death. The GA population is used to replace the SA combinatoric representation inside states, where a population has a fixed size. When used for feature selection, the genetic algorithm individuals are coded in a way that will allow representing the presence or absence of a feature on a feature vector $F$, as explored by \cite{babatunde2014genetic}. By using the proposed coding method, each individual feature vector from the population $P$ is equals $I = \{i_1, ..., i_n\}$, where each $i_i \in I$ is $0$ if the feature $i$ is absent and $1$ otherwise. 

The algorithm in Fig. \ref{alg:gen}, receives an evaluation function $g$, a population size $popSize$, a feature set size $featureSize$ and an iteration limit $t$. It starts by initializing the population $P$, composed of $fpopSize$ individuals coded as the vector $I$, where $featureSize$ is the total amount of features. Next, it will perform the GA routine until $t$ reaches $0$, where a population from each possible value of $t$ is called generation. The main routine of the GA is performed by the function $select$, used to select individuals from $P$ that will be combined in order to generate two new individuals, $breed$, used to combine two individuals, and $mutate$, used to randomly change the value of an individual according to a threshold.

\begin{algorithm}[t]
	\SetKwInOut{Input}{input}
	\SetKwInOut{Output}{output}
	\Input{Evaluation function $g$, population size $popSize$, feature set size $featureSize$, iteration limit $t$}
	\Output{Selected features set $S$}
	\BlankLine
	\DontPrintSemicolon
	\Begin{
		$P = init(popSize, featureSize)$\;
		
		\While{$t \ge 0$}
		{
			$S \gets select(P)$\;
			$B \gets breed(S)$\;
			$B \gets mutate(B)$\;
			$P \gets B$\;
		}	
	
		$S \gets max(P,g)$\;
		return $S$\;
	}
	
	\vspace{-10px}
	\caption{Simple Genetic Algorithm for feature selection pseudo-code.}
	\label{alg:gen}
	\vspace{-10px}
\end{algorithm} 

When ending performing its routine, the GA will calculate the value of $g$ for each individual inside the last generated $P$. The highest individual inside $P$ that is identified by computing $g$ is assigned as $S$, thus allowing it to be used it to select features according to the feature selection model described for a GA. The GA can handle well the feature selection process, but it also relies on a higher computational time, since it is tied to the classifier and to its internal structure of \textit{selecting}, \textit{breeding}, and \textit{mutating}.

%% file: tex/5_conclusion.tex
\section{Conclusion}

Feature selection and extraction is a way to reduce data dimensionality and facilitate the recognition of objects. Although all the feature selection methods aim in improving the identification of relevant data, they are applied in different ways, that may have a high or low cost depending on the problem. \textit{Filter} approaches act as a pre-processing step and do not rely on a classifier\textquotesingle s response. \textit{Filter} methods are considered inefficient to select information when compared to \textit{Wrapper} methods. By contrast, \textit{Wrappers} selects features by interacting with a classifier, which may have better accuracy in identifying features. However, \textit{Wrapped} methods may not be recommended for high dimensional data. An \textit{Embedded} method may be an interesting way in getting advantages of both, \textit{Filters} and \textit{Wrapper} methods since it is a hybrid approach. Therefore, the presented feature selection methods can be accurate and efficient if applied considering their positives and negatives, finding which one fits best the problem\textquotesingle s domain may be the hardest task.